\documentclass[times, review, 10pt]{elsarticle}

\usepackage{amssymb}
\usepackage{amsmath}
\usepackage{graphicx}
\usepackage{multirow}
\usepackage{pifont}

\newdefinition{rmk}{Remark}
\newproof{pf}{Proof}
\newproof{pot}{Proof of Theorem \ref{thm2}}

\journal{Pattern Recognition}

\begin{document}

\begin{frontmatter}



\title{SGANet: Semantic and Geometric Alignment for Multimodal Multi-view Anomaly Detection}

\author[1]{Letian Bai\fnref{label1}}
\ead{lbai799@connect.hkust-gz.edu.cn}

\author[2]{Chengyu Tao\fnref{label2}}
\ead{taochengyu@hnu.edu.cn}

\author[1,3]{Juan Du\corref{cor1}\fnref{label3}}
\ead{juandu@ust.hk}

\fntext[label1]{Writing – original draft, Visualization, Validation, Software, Methodology, Formal analysis, Conceptualization}
\fntext[label2]{Writing – review \& editing, Formal analysis, Data curation}
\fntext[label3]{Writing – review \& editing, Supervision, Project administration}

% \author[1]{Letian Bai}
% \ead{lbai799@connect.hkust-gz.edu.cn}

% \author[2]{Chengyu Tao}
% \ead{taochengyu@hnu.edu.cn}

% \author[1,3]{Juan Du\corref{cor1}}
% \ead{juandu@ust.hk}

\affiliation[1]{organization={Smart Manufacturing Thrust, The Hong Kong University of Science and Technology (Guangzhou)},
                city={Guangzhou},
                postcode={511453}, 
                country={China}}

\affiliation[2]{organization={College of Mechanical and Vehicle Engineering, Hunan University},
                city={Changsha},
                postcode={410082}, 
                country={China}}

\affiliation[3]{organization={The Hong Kong University of Science and Technology},
                city={Hong Kong SAR},
                postcode={999077}, 
                country={China}}

\cortext[cor1]{Corresponding author}

%% Abstract
\begin{abstract}
Multi-view anomaly detection aims to identify surface defects on complex objects using observations captured from multiple viewpoints. However, existing unsupervised methods often suffer from feature inconsistency arising from viewpoint variations and modality discrepancies.  
To address these challenges, we propose a Semantic and Geometric Alignment Network (SGANet), a unified framework for multimodal multi-view anomaly detection that effectively combines semantic and geometric alignment to learn physically coherent feature representations across viewpoints and modalities.
SGANet consists of three key components. The Selective Cross-view Feature Refinement Module (SCFRM) selectively aggregates informative patch features from adjacent views to enhance cross-view feature interaction. The Semantic-Structural Patch Alignment (SSPA) enforces semantic alignment across modalities while maintaining structural consistency under viewpoint transformations. The Multi-View Geometric Alignment (MVGA) further aligns geometrically corresponding patches across viewpoints.
By jointly modeling feature interaction, semantic and structural consistency, and global geometric correspondence, SGANet effectively enhances anomaly detection performance  in multimodal multi-view settings.
Extensive experiments on the SiM3D and Eyecandies datasets demonstrate that SGANet achieves state-of-the-art performance in both anomaly detection and localization, validating its effectiveness in realistic industrial scenarios.
\end{abstract}


% %Research highlights
% \begin{highlights}
% \item SGANet: An unsupervised multimodal multi-view framework for anomaly detection.
% \item Feature alignment combining semantic and geometric strategies for robust detection.
% \item SOTA on SiM3D and Eyecandies for anomaly detection and localization.
% \end{highlights}

%% Keywords

\end{frontmatter}

\section{Introduction}
Industrial anomaly detection (AD) plays a critical role in manufacturing, as subtle surface defects can significantly compromise product quality and safety \cite{cao2024surveyvisualanomalydetection,liang20253d, du20253d}. 
Due to the scarcity of defective samples in real-world industrial scenarios, unsupervised anomaly detection methods trained solely on normal (anomaly-free) samples have become widely adopted for industrial inspection.
However, most existing unsupervised AD approaches rely on single-view observations, which are often insufficient for inspecting complex industrial objects. 
In practice, occlusions and complex 3D geometries prevent a single viewpoint from fully capturing all surface regions of an object. 
Consequently, multi-view observation setups that capture objects from different viewpoints have emerged as an effective paradigm for comprehensive industrial inspection, motivating increasing interest in multi-view anomaly detection \cite{9810850}.

\begin{figure*}
\centering
\includegraphics[width=1.\textwidth]{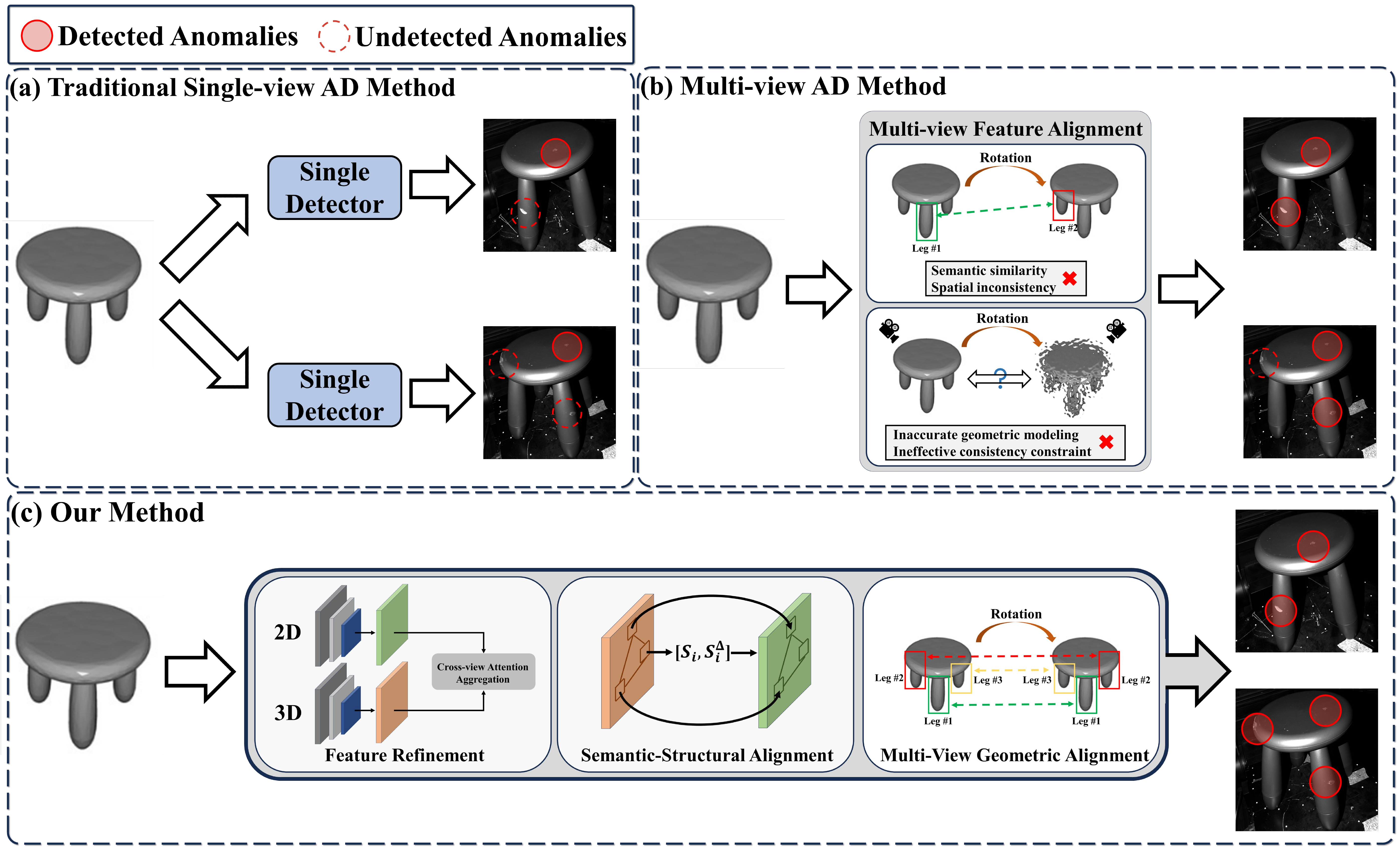}
\caption{    
Comparison of anomaly detection methods. Note that red solid circles denote correctly detected anomalies, while red dashed circles indicate undetected anomalies.
(a) Traditional single-view anomaly detection methods process each view independently, making it difficult to leverage information from other viewpoints.
(b) Multi-view anomaly detection methods integrate information across viewpoints via feature alignment. However, existing approaches suffer from semantic ambiguity due to reliance on similarity without geometric properties, inaccurate spatial modeling from approximate geometric correspondence, and the lack of effective consistency constraints.
(c) Our method explicitly models relationships across views through feature refinement, semantic-structural alignment, and multi-view geometric alignment, enabling effective integration of complementary information and more reliable anomaly detection under viewpoint variations.
}
\label{Fig: Single-view vs multi-view}
\end{figure*}

In practical industrial inspection, many existing anomaly detection approaches are limited to either a single modality or a single viewpoint, which restricts their ability to comprehensively capture complex defect patterns. 
Single-modality methods, such as image-based approaches \cite{PatchCore, PaDiM, FANG2020107474,YANG2022108874, ZAVRTANIK2021107706, NIE2026113261} and point cloud-based approaches \cite{IAENet,3D-ST,LI2025103356}, inevitably suffer from incomplete observations, as appearance and geometric cues often provide complementary information. 
Similarly, traditional anomaly detection frameworks, including reconstruction-based approaches (e.g., DRAEM \cite{DRAEM}, DSR \cite{DSR}, Anomalydiffusion \cite{Anomalydiffusion}) and embedding-based approaches (e.g., CFlow-AD \cite{CFLOW-AD}, PNI \cite{PNI}, LSFA \cite{LSFA}), often struggle to handle occlusions and appearance variations caused by viewpoint changes.
As illustrated in Fig. \ref{Fig: Single-view vs multi-view}(a), traditional single-view methods process each view independently and cannot leverage information across different viewpoints. 
Consequently, defects that are visible only from specific perspectives can be easily overlooked.

Recent studies have therefore begun to explore the integration of information beyond a single observation, incorporating either complementary modalities \cite{M3DM,CFM, 2M3DF, G2SF} or multiple viewpoints \cite{MVAD, MVEAD, yu2025learningmultiviewmulticlassanomaly, Mao_Lian_Wang_Liu_Zheng_Wei_2025}. 
In particular, multi-view anomaly detection methods leverage observations from different viewpoints to provide more comprehensive information, enabling the detection of defects that may be invisible in single views.
In industrial settings, collecting large-scale nominal datasets is costly and time-consuming, and may introduce annotation errors that degrade data quality \cite{SiM3D}. 
Consequently, anomaly detection methods are expected to maintain reliable performance under limited-data conditions, imposing stricter requirements on accuracy and robustness. 
Multi-view observations help alleviate this issue by providing diverse appearance and geometric cues from different viewpoints. 
However, effectively leveraging such complementary information requires learning coherent feature representations across views, which is essential for reliable anomaly detection.
Moreover, most existing multi-view methods are primarily image-based, making it difficult to explicitly capture geometric relationships across views. 
As a result, they rely on indirect or coarse geometric supervision and fail to establish precise cross-view correspondences, limiting the coherence of learned representations. 
In contrast, the multimodal setting provides access to 3D information, enabling features from different viewpoints to be associated with consistent physical locations and thereby facilitating more accurate cross-view alignment.
These advances have stimulated increasing research efforts to extend traditional anomaly detection frameworks to multimodal multi-view settings.

Despite recent progress, learning cross-view relationships remains fundamentally challenging.
One fundamental challenge is that observations from multiple viewpoints are inherently incomplete and distributed. 
Each view captures only a partial projection of the underlying object, resulting in fragmented information across viewpoints. 
Effectively integrating these heterogeneous observations into a unified and informative representation is fundamentally challenging.
On the one hand, semantic ambiguity arises when methods rely solely on embedding similarity without considering geometric properties, leading to incorrect correspondences between semantically similar but spatially distinct regions.
For example, different legs of a plastic stool may share similar semantic features but correspond to different spatial locations (Fig. \ref{Fig: Single-view vs multi-view}(b)).
On the other hand, inaccurate geometric modeling limits the effectiveness of geometry-based approaches.
Methods based on geometric modeling, such as epipolar geometry or homography transformations, leverage geometric properties to guide feature interactions. 
However, these relationships rely on 2D approximations and fail to capture accurate spatial correspondences in the underlying physical structure.
Moreover, enforcing consistency constraints across different views and modalities is inherently challenging in multi-view scenarios.
Due to viewpoint variations and modality discrepancies, features corresponding to the same physical region may undergo significant changes,  
while observations from different modalities can exhibit substantial discrepancies even when capturing the same underlying structure.
Such inconsistency arises from the mismatch between observation space and the underlying physical structure. 
In the absence of effective constraints, these variations can lead to unstable and incoherent feature representations, making it difficult for the model to capture reliable and physically meaningful patterns.
These challenges collectively hinder the learning of physically consistent representations across views. As a result, a fundamental problem in multi-view anomaly detection is how to establish reliable cross-view correspondences that are both semantically meaningful and geometrically consistent in few-shot scenarios.

Motivated by these challenges, we propose SGANet, a unified framework for multimodal multi-view anomaly detection (Fig. \ref{Fig: Single-view vs multi-view}(c)) that learns spatially consistent representations across views and modalities.
Specifically, we leverage semantic cues across modalities and adjacent views to guide cross-view feature refinement while introducing semantic–structural alignment to regularize the learning process, enabling informative aggregation and preventing ambiguous correspondences caused by relying solely on semantic similarity.
To incorporate geometric modeling and effective consistency constraints, we integrate multi-view geometric alignment based on spatial correspondence with semantic–structural alignment to jointly enforce multi-level consistency, thereby learning physically consistent cross-view representations across viewpoints and modalities.
Extensive experiments on the SiM3D \cite{SiM3D} and Eyecandies \cite{Eyecandies} datasets demonstrate that SGANet consistently outperforms state-of-the-art baselines in both anomaly detection and localization tasks.

The contributions of this paper are summarized as follows:

\begin{itemize}
\item We propose SGANet, an unsupervised framework for multimodal multi-view anomaly detection that learns unified representations across viewpoints and modalities.
\item We propose a unified feature alignment framework that addresses semantic ambiguity and the lack of effective consistency constraints while incorporating geometric correspondence to improve cross-view alignment, by jointly modeling semantic, structural, and geometric relationships across viewpoints and modalities through the Selective Cross-View Feature Refinement Module (SCFRM), the Semantic–Structural Patch Alignment (SSPA), and the Multi-View Geometric Alignment (MVGA).
\item Extensive experiments on the SiM3D and Eyecandies datasets demonstrate that SGANet achieves state-of-the-art performance in both detection and localization tasks under real-world industrial inspection scenarios.
\end{itemize}

\section{Literature Review}

\subsection{Multimodal Anomaly Detection}
Early anomaly detection methods predominantly rely on unimodal inputs, such as RGB images or 3D point clouds. 
Image-based approaches are effective at detecting texture and color defects but are insensitive to geometric deviations.
In contrast, 3D-based methods excel at capturing geometric anomalies but often struggle to detect texture and color variations.
This modality gap limits their ability to detect complex defects in real-world scenarios.

To address this limitation, multimodal anomaly detection methods have been proposed to jointly exploit complementary appearance and geometric information.
MMRD \cite{MMRD} utilizes a multimodal reverse distillation approach with a siamese teacher network to extract features from RGB images and depth maps. 
M3DM \cite{M3DM} leverages mutual information to promote cross-modal feature fusion and computes anomaly scores through a one-class SVM. 
M3DM-NR \cite{M3DM-NR} further extends M3DM with a two-stage denoising network and an aligned multi-scale point cloud feature extraction module, replacing farthest point sampling (FPS). 
Meanwhile, CFM \cite{CFM} learns cross-modal mappings for feature alignment, and 2M3DF \cite{2M3DF} improves cross-modal alignment by rendering multi-view RGB images to learn feature correspondences.

Beyond feature fusion, effective score aggregation strategies have also been explored, as multimodal anomaly scores often capture complementary information. 
BTF \cite{BTF} concatenates 2D and 3D features within the PatchCore \cite{PatchCore} framework, which can be regarded as a form of linear score fusion. 
Shape-Guided \cite{Shape-Guided} represents local 3D patches using signed distance functions (SDF) and constructs SDF-guided memory banks for anomaly detection, where anomaly scores from RGB images and SDF representations are combined through a maximum operation.
G$^2$SF \cite{G2SF} further improves score fusion by learning anisotropic local distance metrics through geometry-guided scaling, enabling more discriminative separation between normal and anomalous patterns.

Despite these advances, most multimodal anomaly detection methods process multi-view inputs independently across views. Consequently, they lack a unified representation across viewpoints and do not explicitly account for geometric correspondence, which may lead to feature misalignment and degraded anomaly detection performance.

\subsection{Multi-view Anomaly Detection}
Compared to traditional anomaly detection, multi-view anomaly detection is an emerging research direction that poses greater challenges due to the need for modeling cross-view relationships and geometric consistency \cite{A_Systematic_Exploration-TNNLS}.

MVAD \cite{MVAD} first introduces a Multi-View Adaptive Selection mechanism that computes local correlations across views and adaptively fuses the most relevant features to enhance feature representations. Although MVAD performs adaptive fusion based on semantic correlations, it does not incorporate explicit geometric properties.
To address this limitation, MVEAD \cite{MVEAD} incorporates epipolar geometry to guide feature fusion. However, its geometric modeling does not explicitly capture the consistency of viewpoint transformations, which is insufficient to ensure accurate feature alignment across multiple views.
VSAD \cite{VSAD-AAAI} proposes a multi-view anomaly detection framework that integrates feature alignment based on homographic properties into a latent diffusion model. 
CPMF \cite{CPMF} constructs complementary pseudo-multimodal features by combining local geometric descriptors derived from point clouds with semantic features extracted from multi-view projections and aggregates them for 3D anomaly detection.
Although leveraging multi-view observations, both VSAD and CPMF are fundamentally designed for single-modality settings.

Despite these advances, existing multi-view anomaly detection methods still exhibit several fundamental limitations. 
One limitation is that existing approaches operate on a single modality, preventing them from fully exploiting the complementary information between appearance and geometry. 
Another limitation lies in the multi-view modeling strategy.  
Existing methods either rely on semantic modeling without explicit geometric modeling or incorporate geometric cues that are only approximated and fail to accurately capture true spatial correspondences. 
Moreover, they lack effective consistency constraints to regulate feature transformations under viewpoint variations. 
To address these limitations, we propose a unified framework for multimodal multi-view anomaly detection that jointly models semantic–structural relationships and geometric correspondence across both views and modalities, enabling the learning of spatially consistent representations from complementary appearance and geometric cues.
In the following section, we present the proposed framework and describe the key components in detail.

\begin{figure*}
\centering
\includegraphics[width=1.\textwidth]{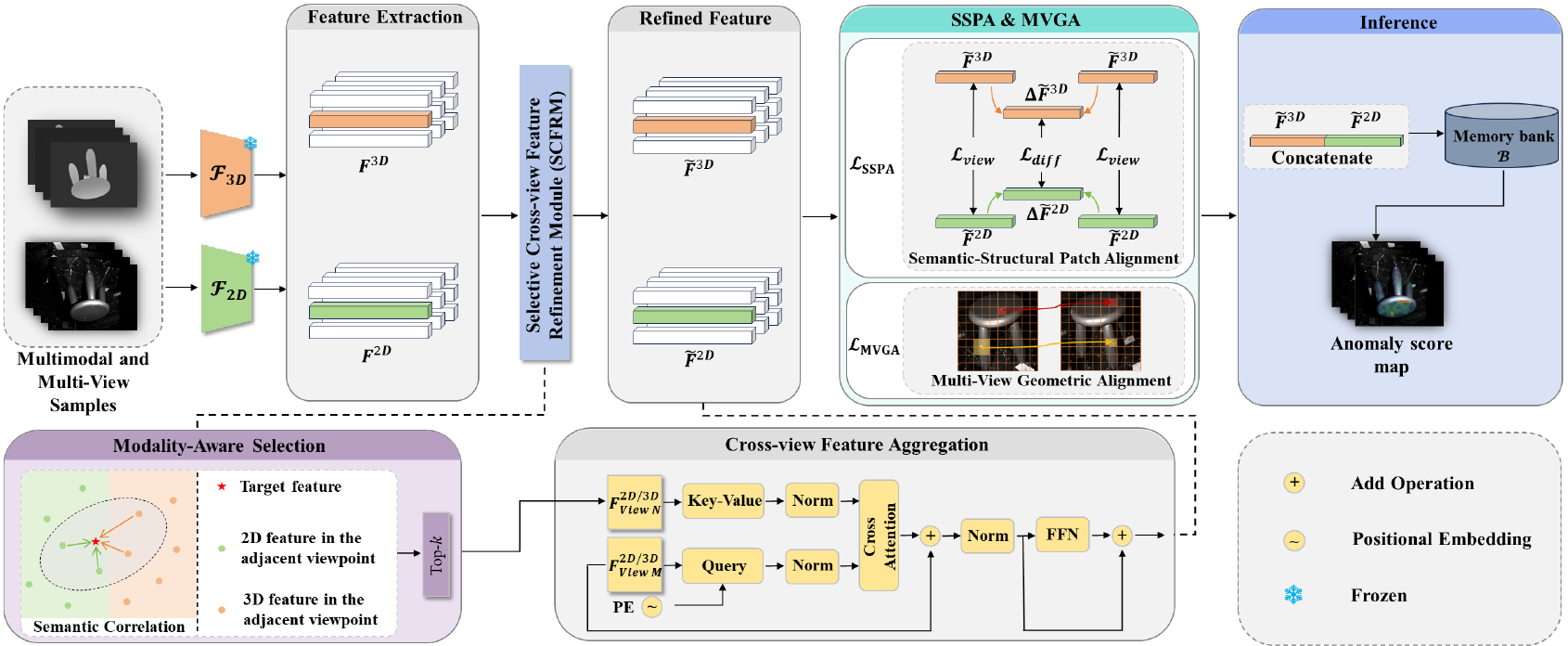}
\caption{
Overview of the proposed SGANet framework for multimodal multi-view anomaly detection. 
Multi-view images and depth maps are first encoded into patch-level features using frozen feature extractors $\mathcal{F}^{2D}$ and $\mathcal{F}^{3D}$, respectively.
The Selective Cross-view Feature Refinement Module (SCFRM, Section \ref{subsec:SCFRM}) first refines feature representations by aggregating semantically related features from adjacent viewpoints through modality-aware selection and cross-view feature aggregation.
Based on the refined features, the Semantic-Structural Patch Alignment (SSPA, Section \ref{subsec:SSPA}) enforces semantic alignment by matching corresponding patches between the 2D and 3D modalities within each view and further incorporates a structural alignment mechanism to maintain consistency under viewpoint transitions.
To further ensure feature coherence, Multi-View Geometric Alignment (MVGA, Section \ref{subsec:MVGA}) leverages explicit geometric correspondences to align features at spatially corresponding locations, thereby reducing feature discrepancies and promoting spatially consistent multi-view feature representations.
During inference (Section \ref{subsec:Learning and Anomaly Scoring}), multimodal features are concatenated and compared with a memory bank constructed from normal samples to compute anomaly scores and generate anomaly maps.}
\label{Fig: overall structure}
\end{figure*}

\section{Methodology}

\subsection{Problem Formulation}\label{subsec:Problem Formulation}
In unsupervised anomaly detection, we are given a training set $\mathcal{X}_{train}$ and a test set $\mathcal{X}_{test}$. 
The training set contains only normal (anomaly-free) samples, while the test set contains both normal and anomalous samples.
In the few-shot setting, the training set contains only a limited number of samples.
In the multimodal multi-view setting considered in this work, each sample $x \in \{\mathcal{X}_{train}, \mathcal{X}_{test}\}$ consists of multiple observations captured from $I$ viewpoints and complementary modalities $\mathcal{M} = \{2D, 3D\}$.
The 2D modality observations correspond to RGB or grayscale images, while the 3D modality observations correspond to depth maps.
A sample can therefore be represented as $x = \{x_i^{m}\}_{i=1,\, m \in \mathcal{M}}^{I}$, where $x_i^{m} \in \mathbb{R}^{C_m \times H \times W}$. 
Here, $m$ denotes the modality index, $C_m$ represents the number of channels for modality $m$, and $H$ and $W$ denote the spatial height and width of each observation. 
The objective of anomaly detection is to learn representations of normal patterns from $\mathcal{X}_{train}$ and detect anomalous deviations in test samples $x \in \mathcal{X}_{test}$ during inference by predicting an anomaly map for each view, along with an overall anomaly score for the sample.

\subsection{Overall Structure}
We propose SGANet, a unified framework for multimodal multi-view anomaly detection that integrates semantic and geometric alignment to learn physically consistent feature representations across different viewpoints and modalities. 
As illustrated in Fig. \ref{Fig: overall structure}, SGANet progressively refines and aligns feature representations through the proposed components, given the multimodal and multi-view samples defined in Section \ref{subsec:Problem Formulation}.
SCFRM first performs cross-view feature refinement to capture informative interactions from different viewpoints and modalities.      
To address the challenge of inconsistent feature representations across viewpoints and modalities, SSPA enforces semantic and structural consistency based on the refined features, aligning cross-modal representations while preserving viewpoint-induced structural variations. 
Furthermore, to explicitly model geometric correspondence, MVGA introduces geometric alignment to promote spatially consistent and physically meaningful representations across views.
Detailed descriptions of each component are presented in Sections \ref{subsec:SCFRM}-\ref{subsec:MVGA}.

\subsection{Selective Cross-View Feature Refinement Module (SCFRM)}\label{subsec:SCFRM}
Multi-view anomaly detection requires learning consistent representations across viewpoints and modalities.
Achieving such consistency typically requires refining features using semantically corresponding patches. 
However, directly aggregating features from all variations and viewpoints may introduce redundant and even noisy information, which can degrade representation quality and increase computational costs.
To address these challenges, we propose the Selective Cross-view Feature Refinement Module (SCFRM), which explicitly models interactions across viewpoints and modalities to refine feature representations.

\textbf{Feature Extraction.}
Our framework takes the multimodal multi-view observations as input. 
For each view $i$ and modality $m \in \mathcal{M}$, we employ pre-trained feature extractors $\mathcal{F}^{2D}$ and $\mathcal{F}^{3D}$ to encode the corresponding 2D and 3D observations into patch-level feature representations. 
The extracted feature map is denoted as $F_i^{m} \in \mathbb{R}^{P \times d_m}$, where $P$ is the number of patches and $d_m$ is the feature dimension of modality $m$. 
The feature at location $p$ in view $i$ is denoted by $f_{i,p}^{m} \in \mathbb{R}^{d_m}$, where $p \in \{1,\dots,P\}$.

\textbf{Modality-Aware Selection.}
To ensure stable cross-view and cross-modal interaction, similarity computation is performed on adjacent viewpoints in the ordered multi-view capture sequence. 
Specifically, for a query feature $f_{i,p}^{m}$ and a candidate feature $f_{j,q}^{m}$ from an adjacent view $j$, we define the modality-aware similarity as
\begin{equation}
\operatorname{sim}^{m}_{i \rightarrow j}(p,q)
=
\alpha \,\operatorname{sim}\bigl(f_{i,p}^m, f_{j,q}^m\bigr)+ (1-\alpha)\,\operatorname{sim}\bigl(f_{i,p}^{\bar{m}}, f_{j,q}^{\bar{m}}\bigr),
\end{equation}
where $\operatorname{sim}(\cdot,\cdot)$ denotes cosine similarity, $\bar m$ denotes the complementary modality of $m$, and the coefficient $\alpha$ balances the contributions of the similarity interaction across modalities. 
In our implementation, we set $\alpha = 0.8$.
This formulation incorporates cross-modal interaction by jointly considering similarities from both modalities, enabling more informative matching across views.
For each query feature in view $i$, we select the top-$k$ most relevant candidate feature indices in the adjacent view $j$ as
\begin{equation}
\mathcal{N}^{m}_{i \rightarrow j}(p)
=
\operatorname{Top\text{-}k}_q
\left(
\operatorname{sim}^{m}_{i \rightarrow j}(p,q)
\right),
\end{equation}
where $\operatorname{Top\text{-}k}_q$ operation selects the $k$ patches with the highest similarity scores with respect to $q$. 
This strategy filters out less relevant or noisy features while retaining the most informative candidates for effective cross-view aggregation.
The final selected candidate set for cross-view interaction is obtained by aggregating the selected feature indices from all adjacent views:
\begin{equation}
\mathcal{N}_{i,p}^{m}
=
\bigcup_{j}
\mathcal{N}^{m}_{i \rightarrow j}(p),
\end{equation}
which contains $2k$ candidate feature pair indices for further cross-view feature aggregation.

\textbf{Cross-View Feature Aggregation.}
Based on the selected candidate set, SCFRM performs cross-view attention to aggregate complementary information from adjacent views. 
For a query feature $f_{i,p}^{m}$, cross-view attention is computed over the candidate features pair indices $(j,q)\in\mathcal{N}_{i,p}^{m}$, where the query, key, and value embeddings are defined as
\begin{equation}
\label{eq:Attention-1}
q_{i,p}^{m} = W_q f_{i,p}^{m}, ~k_{j,q}^{m} = W_k f_{j,q}^{m}, ~v_{j,q}^{m} = W_v f_{j,q}^{m},
\end{equation}
with $W_q$, $W_k$, and $W_v$ denoting learnable projection matrices.
The attention weights are computed based on the attention mechanism:
\begin{equation}
\label{eq:Attention-2}
\alpha_{j,q}^{m}
=
\frac{
\exp \left( (q_{i,p}^{m})^\top k_{j,q}^{m} / \sqrt{d_m} \right)
}{
\sum\limits_{(j',q') \in \mathcal{N}_{i,p}^{m}}
\exp \left( (q_{i,p}^{m})^\top k_{j',q'}^{m} / \sqrt{d_m} \right)
}.
\end{equation}

Based on the attention mechanism, the refined feature representation can be obtained as
\begin{equation}
\label{eq:Refined feature}
\tilde{f}_{i,p}^{m}
=
\sum_{(j,q) \in \mathcal{N}_{i,p}^{m}}
\alpha_{j,q}^{m} \, v_{j,q}^{m}.
\end{equation}

By performing cross-view interaction on adjacent viewpoints and selecting the top-$k$ relevant patches based on modality-aware selection, SCFRM performs selective cross-view aggregation that suppresses weakly correlated or noisy responses.
The attention mechanism is applied to each feature for each view and modality, allowing for the aggregation of complementary information from semantically corresponding features across adjacent views.
The refined features provide stable representations for subsequent semantic and structural consistency patch alignment while maintaining computational efficiency.

\subsection{Semantic-Structural Patch Alignment (SSPA)} \label{subsec:SSPA}
While SCFRM refines feature representations based on adjacent views, robust multimodal anomaly detection further requires feature alignment across viewpoints and modalities. 
In particular, features corresponding to the same location across modalities should be aligned to maintain semantic consistency, while feature differences across viewpoints should be constrained to capture meaningful structural variations.
To address these requirements, we introduce the Semantic-Structural Patch Alignment (SSPA), which enforces semantic and structural consistency through contrastive learning across modalities.

Let $\tilde{F}_i^{m} = \{\tilde{f}_{i,p}^{m}\}_{p=1}^{P} \in \mathbb{R}^{P \times d_m}$ denote the feature map obtained after refinement by SCFRM.
Specifically, we compute the semantic similarity matrix between the two modalities as
\begin{equation}
S_i=\tilde{F}_i^{2D}\left(\tilde{F}_i^{3D}\right)^\top.
\end{equation}

The semantic consistency alignment loss for view $i$ is formulated using the InfoNCE loss \cite{InfoNCE}:
\begin{equation}
\mathcal{L}^{(i)}_{\mathrm{view}}=-\frac{1}{P}\sum_{p=1}^{P}\log
\frac{\exp\left(S_i(p,p)\right)}{\sum_{q=1}^{P}\exp\left(S_i(p,q)\right)}, 
\end{equation}
where $S_i(p,p)$ denotes the similarity between corresponding cross-modal features at location $p$, forming the positive pair, while $S_i(p,q)$ measures the similarities between the query feature $p$ and all candidate features $q$ in the other modality.

While semantic patch alignment enforces semantic consistency within each view, it does not explicitly capture structural variations across viewpoints. 
Since viewpoint transformations primarily change the observation perspective rather than the underlying structure, the resulting structural variations should remain spatially consistent across modalities.
The difference between features from adjacent views reflects the structural changes caused by viewpoint transformation, and enforcing structural consistency encourages the model to learn coherent feature representations.
To model such variations, we introduce differential features computed between adjacent views.
Specifically, differential features are constructed between two adjacent views as
\begin{equation}
\Delta \tilde{F}^{\mathrm{2D}}_{i}=\tilde{F}^{\mathrm{2D}}_{i+1}-\tilde{F}^{\mathrm{2D}}_i,
~\Delta \tilde{F}^{\mathrm{3D}}_{i}=\tilde{F}^{\mathrm{3D}}_{i+1}-\tilde{F}^{\mathrm{3D}}_i.
\end{equation}

The corresponding structural similarity matrix is
\begin{equation}
S^{\Delta}_i=\Delta \tilde{F}^{\mathrm{2D}}_{i}\left(\Delta \tilde{F}^{\mathrm{3D}}_{i}\right)^\top,
\end{equation}
and the structural consistency alignment loss for view $i$ is formulated as
\begin{equation}
\mathcal{L}^{(i)}_{\mathrm{diff}}=-\frac{1}{P}\sum_{p=1}^{P}\log
\frac{\exp\left(S^{\Delta}_i(p,p)\right)}{\sum_{q=1}^{P}\exp\left(S^{\Delta}_i(p,q)\right)},
\end{equation}
where $S^{\Delta}_i(p,p)$ denotes the similarity between corresponding differential features across modalities at location $p$, forming the positive pair, while $S^{\Delta}_i(p,q)$ measures the similarities between the differential feature at location $p$ and all candidate features $q$ in the other modality.
Therefore, the overall consistency alignment loss is given by
\begin{equation}
\mathcal{L}_{\mathrm{view}}=\frac{1}{I}\sum_{i=1}^{I}\mathcal{L}^{(i)}_{\mathrm{view}},
~\mathcal{L}_{\mathrm{diff}}=\frac{1}{I-1}\sum_{i=1}^{I-1}\mathcal{L}^{(i)}_{\mathrm{diff}},
\end{equation}
and the final objective is therefore given by
\begin{equation}
\mathcal{L}_{\mathrm{SSPA}}=\mathcal{L}_{\mathrm{view}}+\mathcal{L}_{\mathrm{diff}}.
\end{equation}

Through these objectives, SSPA enforces semantic alignment across modalities with $\mathcal{L}_{\mathrm{view}}$ and structural alignment under viewpoint transformations with $\mathcal{L}_{\mathrm{diff}}$.
To further incorporate geometric relationships among different views, we introduce the Multi-View Geometric Alignment (MVGA) strategy.

% \textcolor{red}{Geometric alignment across views, geometric coherence.}
\subsection{Multi-View Geometric Alignment (MVGA)}\label{subsec:MVGA}
While SSPA focuses on enforcing semantic and structural consistency across viewpoints and modalities, multi-view anomaly detection also requires geometric coherence corresponding to the same spatial location.

\textbf{Global Geometric Correspondence.}
In industrial multi-view capture setups with calibrated cameras, stable geometric correspondence exists across viewpoints. 
For a given view $i$, we define its neighboring view set as
\begin{equation}
\mathcal{V}(i) 
= \{ i-N, \dots, i-1, i+1, \dots, i+N \},
\end{equation}
where $N$ controls the number of preceding and succeeding views considered for alignment.

By leveraging the known camera intrinsic and extrinsic parameters, locations in view $i$ can be projected into each neighboring view $j \in \mathcal{V}(i)$ to establish geometric correspondences. 
This procedure yields a set of correspondence pairs $\Omega_{i,j}$ for $i=1,\dots,I$ and $j \in \mathcal{V}(i)$, where each pair $(p,q) \in \Omega_{i,j}$ represents spatially aligned locations at position $p$ in view $i$ and position $q$ in view $j$, corresponding to the same physical point observed from different viewpoints.
Projections that fall outside image boundaries or correspond to occluded regions are excluded from the correspondence pairs set $\Omega_{i,j}$.

\textbf{Geometric Alignment Loss.}
Let $\tilde{\mathbf{f}}^{m}_{i,p}$ and $\tilde{\mathbf{f}}^{m}_{j,q}$ denote the refined features at corresponding locations $(p,q)\in\Omega_{i,j}$ in views $i$ and $j$ for modality $m$.
The pairwise alignment loss between views $i$ and $j$ is defined as
\begin{equation}
\mathcal{L}_{\mathrm{mvga}}^{(i,j)}
=
\frac{1}{|\mathcal{M}|}
\sum_{m \in \mathcal{M}}
\frac{1}{|\Omega_{i,j}|}
\sum_{(p,q)\in \Omega_{i,j}}
\left\|
\tilde{\mathbf{f}}^{m}_{i,p}
-
\tilde{\mathbf{f}}^{m}_{j,q}
\right\|_2, 
\end{equation}
where $|\cdot|$ denotes the cardinality of a set.
This formulation enforces consistency between features at geometrically corresponding locations
The overall geometric alignment loss is defined by aggregating the pairwise losses across all views and their neighboring views:
\begin{equation}
\mathcal{L}_{\mathrm{MVGA}}
=
\frac{1}{I}
\sum_{i=1}^{I}
\frac{1}{|\mathcal{V}(i)|}
\sum_{j\in\mathcal{V}(i)}
\mathcal{L}_{\mathrm{mvga}}^{(i,j)}.
\end{equation}

Through this design, MVGA leverages global geometric correspondences across viewpoints to align features at matched spatial locations, promoting spatially coherent representations across views.
This explicit geometric alignment mitigates feature instability and promotes spatially consistent representations for robust multi-view anomaly detection.

\subsection{Learning and Anomaly Scoring}\label{subsec:Learning and Anomaly Scoring}

During training, the model is optimized using only normal samples to learn discriminative feature representations across viewpoints and modalities. 
Given multi-view observations from the modalities $\mathcal{M}=\{2D,3D\}$, the framework jointly optimizes the objectives of SSPA and MVGA, while SCFRM serves as the feature refinement backbone.
The overall training objective is defined as
\begin{equation}
\mathcal{L} =  \lambda_{\text{SSPA}} \mathcal{L}_{\text{SSPA}} + \lambda_{\text{MVGA}} \mathcal{L}_{\text{MVGA}},
\end{equation}
where $\lambda_{\text{SSPA}}$ and $\lambda_{\text{MVGA}}$ control the contributions of each alignment objective function.

After training, the refined feature maps extracted from normal samples are used to construct a multimodal memory bank $\mathcal{B}$. 
For each view $i$, the refined features from the two modalities are fused through channel-wise concatenation:
\begin{equation}
\tilde F_i = \tilde{F}_i^{2D} \oplus \tilde{F}_i^{3D},
\end{equation}
where $\oplus$ denotes concatenation along the feature dimension, resulting in $\tilde F_i \in \mathbb{R}^{P \times (d_{2D} + d_{3D})}$. 

At inference time, the same feature extraction and refinement pipeline is applied to the test samples.
Following the distance-based anomaly detection paradigm \cite{PatchCore},
the anomaly score for feature $\tilde f_{i,p}$ is computed as the Euclidean distance to its nearest neighbor in the memory bank $\mathcal{B}$:
\begin{equation}
s_{i,p} = \min_{\tilde f_k \in \mathcal{B}} \left\|\tilde f_{i,p} - \tilde f_k \right\|_2.
\end{equation}

Based on the patch-level scores, the view-level anomaly score $s_i$ is obtained by taking the maximum value in the anomaly map of view $i$. 
The sample-level anomaly score $s$ is then computed as the maximum score across all views.
% 需要修改：1.不是sample而是instance；2.应该是像素，而不是instance/sample

\section{Experiment}
\subsection{Experiment Details}
\textbf{Datasets.} Experiments are conducted on the SiM3D \cite{SiM3D} dataset, which jointly supports multimodal and multi-view anomaly detection. 
SiM3D is a real-world industrial dataset comprising eight categories of manufactured objects. 
The dataset contains 331 real samples, where each category provides only a single normal sample for training, while the remaining 14–98 samples are used for testing.
Each object is captured from 12 or 36 viewpoints, enabling systematic evaluation of multimodal multi-view anomaly detection under realistic industrial settings.
Detailed statistics of the SiM3D dataset are provided in Table~\ref{tab:Dataset statistics of the SiM3D dataset}.

In addition to the SiM3D dataset, we also conduct experiments on the Eyecandies dataset \cite{Eyecandies}, a benchmark consisting of 10 object categories.
To maintain a consistent evaluation protocol with SiM3D, we select one sample from the training set and render it along with all test samples to construct a multi-view setting.
Specifically, we extend the Eyecandies dataset using the rendering pipeline of CPMF \cite{CPMF} and PointAD \cite{PointAD}. 
Multiple viewpoints are synthesized by rotating the camera around the object center while keeping the camera intrinsics fixed. 
Rotations are applied along the X-axis with angles of $x \in \{0, \pm \pi/12\}$ and the Y-axis with angles of $\{0, \pm \pi/12, \pi/6\}$.
By combining these rotations, a total of 12 viewpoints are generated for each sample.
For each viewpoint, RGB images, depth maps, and anomaly masks are generated via geometric projection of 3D points using calibrated camera parameters.

\begin{table}[h]
\centering
\caption{Dataset statistics of the SiM3D dataset \cite{SiM3D}.}
\label{tab:Dataset statistics of the SiM3D dataset}
\begin{tabular}{lcccc}
\hline 
Category & Nominal & Anomalous & Total & Views \\ \hline 
Bathroom Furniture & 8 & 11 & 19 & 36 \\
Container & 46 & 47 & 93 & 12 \\
Plastic Stool & 10 & 11 & 21 & 12 \\
Plastic Vase & 48 & 50 & 98 & 12 \\
Rubbish Bin & 20 & 20 & 40 & 12 \\
Sink Cabinet & 9 & 8 & 17 & 36 \\
Wicker Vase & 10 & 11 & 21 & 12 \\
Wooden Stool & 6 & 8 & 14 & 12 \\ \hline
\end{tabular}
\end{table}

\textbf{Evaluation Metrics.}
For object-level anomaly detection, we report the Instance-level Area Under the Receiver Operator Curve (I-AUROC) on both the SiM3D and Eyecandies datasets. For point-level anomaly detection, we report the Voxel-level Area Under the Per-Region Overlap at a 1\% integration limit (V-AUPRO@1\%) on the SiM3D dataset, while the Pixel-level AUROC (P-AUROC) and Area Under the Per-Region Overlap at a 30\% integration limit (AUPRO@30\%) on the Eyecandies dataset.

\textbf{Implementation Details.}
In our implementation, the feature extractors $\mathcal{F}^{2D}$ and $\mathcal{F}^{3D}$ are instantiated using a frozen DINO-v2 backbone \cite{Dino-v2} to extract feature representations.
For grayscale images in SiM3D and depth maps in both SiM3D and Eyecandies, the single-channel inputs are replicated along the channel dimension to form three-channel images, ensuring compatibility with the input format of DINO-v2. 
All images are then resized to $518 \times 518$ for SiM3D and $512 \times 512$ for Eyecandies before being fed into the feature extractor.
For multi-view interaction, the number of neighboring views is set to $N=2$, and the top-$k$ most relevant patches ($k=8$) are selected.
The loss weights are set to $\lambda_{\text{SSPA}}=1$ and $\lambda_{\text{MVGA}}=2$, respectively.
Due to inconsistencies between the dataset statistics reported in the original SiM3D paper and those in the released dataset, we re-evaluated some representative baseline methods using the publicly available implementations. 
Following the official evaluation protocol of SiM3D, the predicted 2D anomaly maps are projected onto the corresponding 3D point clouds for performance evaluation.

\subsection{Comparison on Multimodal Benchmarks}
\textbf{Quantitative Results on SiM3D.}
Tables \ref{tab:SiM3D I-AUROC} and \ref{tab:SiM3D V-AUPRO@1 percent} report the results of I-AUROC and V-AUPRO@1\% on the SiM3D dataset. SGANet achieves the highest mean I-AUROC of 0.887, outperforming all baseline methods. 
For anomaly localization, SGANet also achieves the best overall performance with a mean V-AUPRO@1\% of 0.614, indicating superior localization precision under strict false-positive constraints. 
Compared with RGB-based approaches such as PatchCore as well as multimodal methods including CFM and M3DM, SGANet achieves more reliable performance on geometrically complex objects, such as Plastic Stool and Rubbish Bin. 
Although PatchCore (DINO-v2) shows competitive performance in terms of V-AUPRO@1\%, its mean I-AUROC of 0.867 remains lower compared to SGANet (0.887).

Moreover, the complementary nature of appearance and geometric cues further enhances the discriminative capability of the learned representations.
Compared with the multi-view baseline MVAD, SGANet improves the mean I-AUROC from 0.721 to 0.887 and the mean V-AUPRO@1\% from 0.542 to 0.614 on the SiM3D dataset. 
This is because MVAD aggregates features across views solely based on semantic similarity within a single modality, which limits its ability to leverage cross-modal and global geometric correspondences.
In addition, the SiM3D dataset contains only a single normal sample for training in each category, resulting in a highly limited training setting.
Under such limited supervision, learning stable feature representations becomes crucial. 
The proposed alignment strategy promotes consistency across viewpoints and modalities, enabling the model to capture reliable patterns of normal objects despite the limited number of training samples.
This property further highlights the robustness of SGANet in few-shot industrial inspection scenarios.

\begin{figure}
	\centering
	\includegraphics[width=\columnwidth]{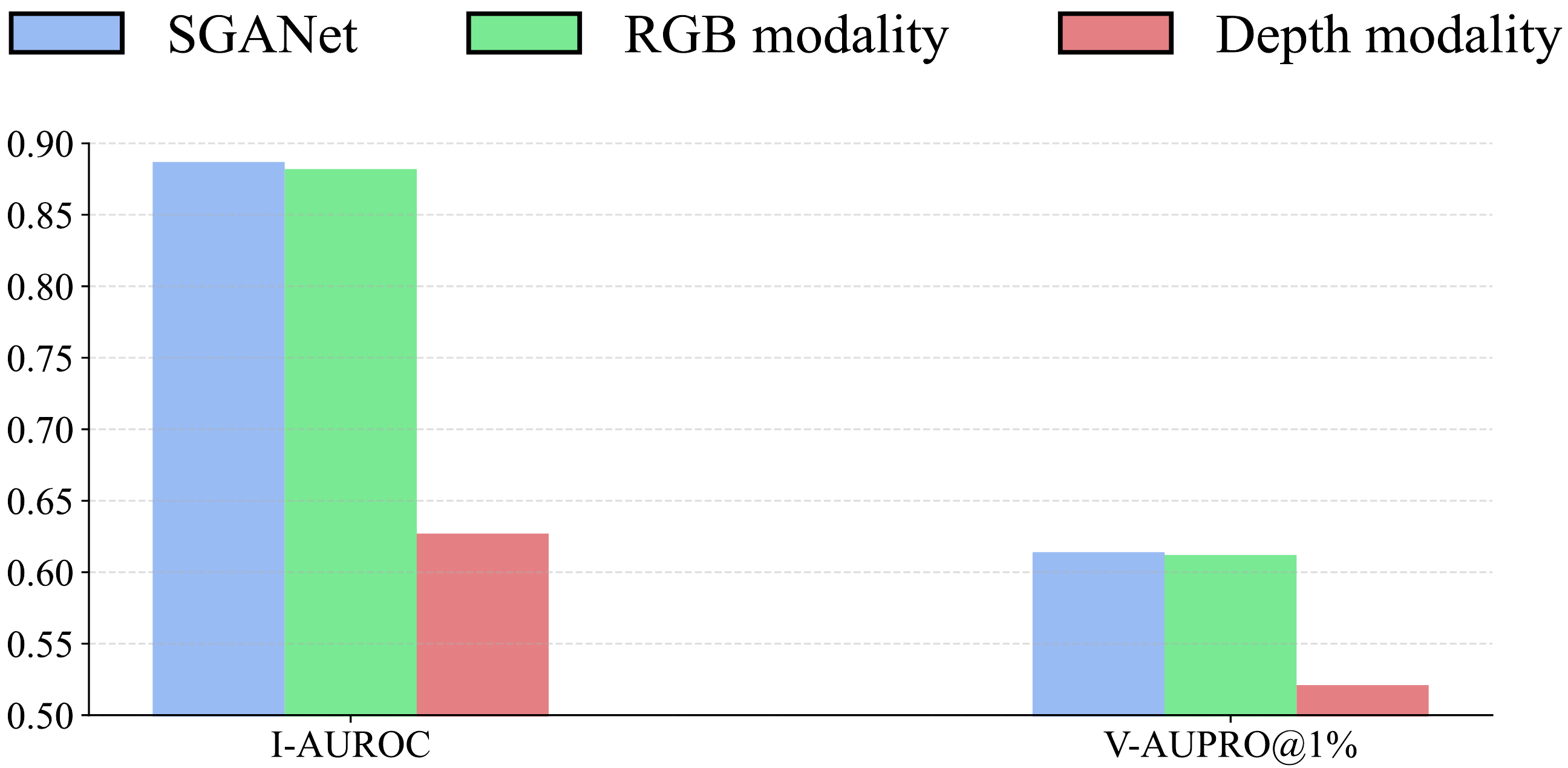}
	\caption{I-AUROC and V-AUPRO@1\% results on the SiM3D dataset using RGB modality, depth modality, and the proposed SGANet framework.}
	\label{Fig: SiM3D Histogram}
\end{figure}

Fig.~\ref{Fig: SiM3D Histogram} presents a comparison between unimodal and multimodal configurations on the SiM3D dataset. 
The multimodal configuration achieves the best performance on both evaluation metrics, with an I-AUROC of 0.887 and a V-AUPRO@1\% of 0.614, compared to 0.882 and 0.612 for the RGB modality and 0.627 and 0.521 for the depth modality. 
The performance gain over the RGB modality is moderate, which can be attributed to the characteristics of the SiM3D dataset.
DINO-v2 is effective in extracting discriminative features from RGB images, whereas depth maps exhibit a noticeable domain gap, limiting the amount of useful information the depth modality can provide.
Nevertheless, the multimodal configuration still leads to improved performance by leveraging the complementary characteristics of appearance and geometric information. 
RGB features primarily capture variations in appearance, whereas depth features provide structural cues. 
By leveraging cross-modal information through the proposed alignment strategy, the model learns more robust complementary representations, leading to superior performance compared with a single modality.

\begin{figure*}
	\centering
	\includegraphics[width=1.\textwidth]{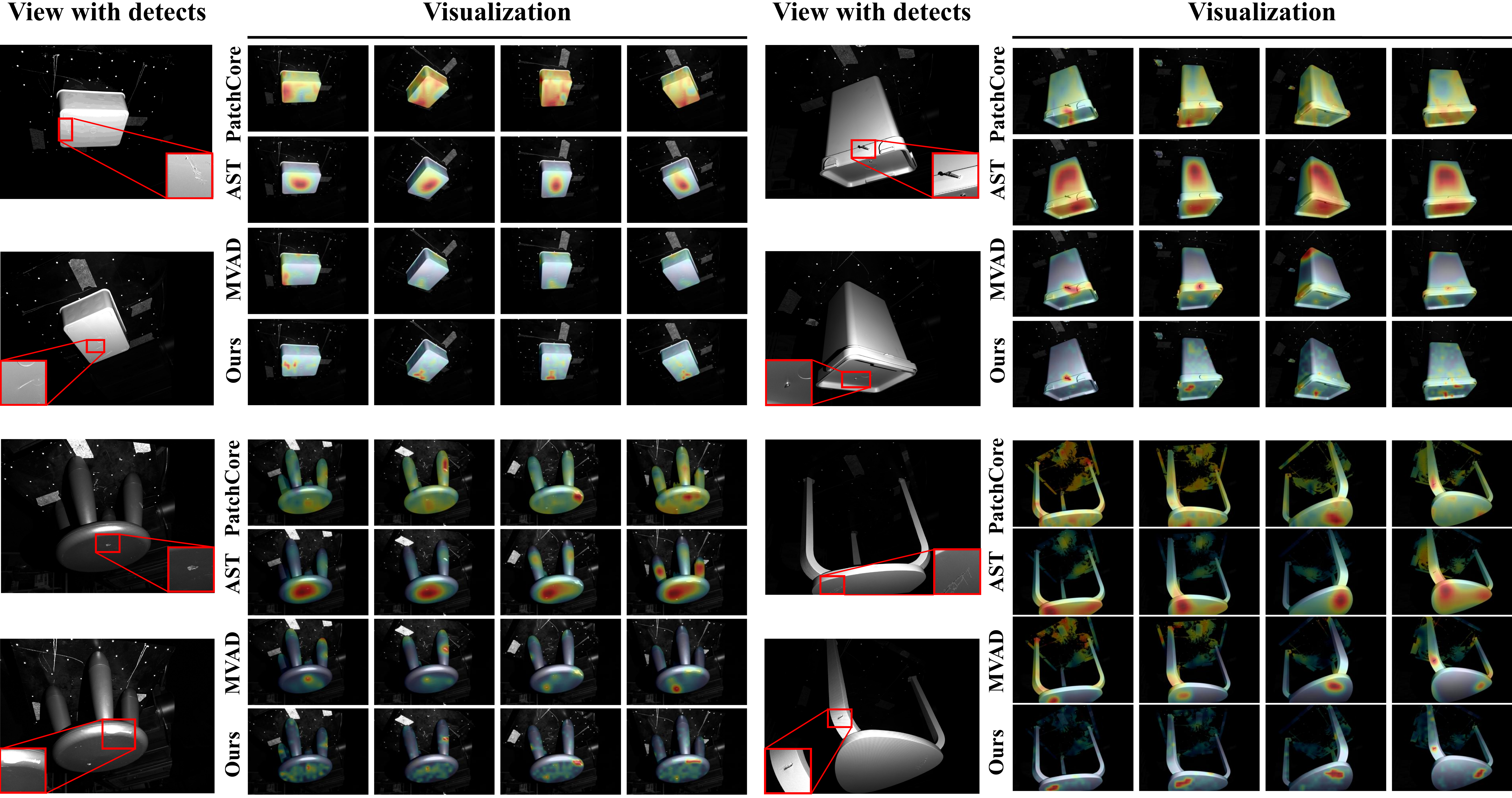}
	\caption{Qualitative comparison on the SiM3D dataset \cite{SiM3D}. Visualizations include defect regions highlighted in the input views and the corresponding anomaly score maps predicted by representative methods, including single-view baselines (PatchCore, AST), the multi-view method (MVAD), and the proposed SGANet framework.}
	\label{Fig: SiM3D Visulization Result}
\end{figure*}

\begin{table*}[ht]
\centering
\caption{I-AUROC result on the SiM3D dataset. Best results in bold, runner-ups \underline{underlined}.}
\label{tab:SiM3D I-AUROC}
\resizebox{\textwidth}{!}{
\begin{tabular}{c c c c c c c c c c c}
\hline 
Method & Modality & Pl. Stool & Rub. Bin & W. Vase & B. Furn. & Cont. & Pl. Vase & W. Stool & Sink Cab. & Mean \\
\hline
PatchCore (WRN-101) \cite{PatchCore}   
& RGB & \underline{0.900} & \textbf{1.000} & \underline{0.836} & 0.795 & 0.753 & 0.553 & 0.917 & \textbf{0.972} & 0.841 \\

PatchCore (DINO-v2) \cite{PatchCore}   
& RGB & 0.736 & \textbf{1.000} & 0.745 & 0.795 & \textbf{0.860} & \textbf{0.865} & \textbf{1.000} & \underline{0.931} &\underline{0.867} \\

EfficientAD \cite{Efficient_AD}           
& RGB & 0.280 & 0.732 & 0.000 & \textbf{0.878} & 0.424 & 0.730 & 0.928 & 0.712 & 0.586 \\

AST \cite{AST}    
& RGB & 0.900 & 0.260 & 0.618 & 0.625 & 0.547 & 0.423 & 0.625 & 0.403 & 0.550 \\

MVAD \cite{MVAD}                 
& RGB & 0.573 & \underline{0.990} & 0.427 & 0.727 & 0.669 & 0.567 & \underline{0.938} & 0.875 & 0.721 \\  

BTF  \cite{BTF}  
& RGB + PC & 0.421 & 0.217 & 0.504 & 0.565 & 0.545 & 0.471 & 0.678 & 0.424 & 0.478 \\

CFM (DINO-v2 + FPFH) \cite{CFM} 
& RGB + PC & 0.373 & 0.313 & 0.018 & 0.523 & 0.330 & 0.488 & 0.750 & 0.611 & 0.426 \\

M3DM (DINO-v2 + FPFH) \cite{M3DM} 
& RGB + PC & 0.702 & 0.988 & 0.661 & 0.545 & 0.556 & 0.649 & 0.392 & 0.475 & 0.621 \\

AST \cite{AST} & RGB + Depth     
& \textbf{1.000} & 0.900 & 0.800 & 0.420 & 0.636 & 0.448 & 0.500 & 0.861 & 0.696 \\

\textbf{SGANet(Ours)} & RGB + Depth  
& 0.845 & \underline{0.990} & \textbf{0.964} & \underline{0.830} & \underline{0.841} & \underline{0.778} & \textbf{1.000} & 0.847 & \textbf{0.887} \\
\hline 
\end{tabular}
}
\end{table*}

\begin{table*}[ht]
\centering
\caption{V-AUPRO@1\% result on the SiM3D dataset. Best results in bold, runner-ups \underline{underlined}.}
\label{tab:SiM3D V-AUPRO@1 percent}
\resizebox{\textwidth}{!}{
\begin{tabular}{c c c c c c c c c c c}
\hline
Method & Modality & Pl. Stool & Rub. Bin & W. Vase & B. Furn. & Cont. & Pl. Vase & W. Stool & Sink Cab. & Mean \\
\hline
PatchCore (WRN-101) \cite{PatchCore} & RGB 
& 0.774 & 0.478 & 0.775 & 0.613 & \underline{0.698} & 0.752 & \textbf{0.290} & 0.431 & 0.601 \\

PatchCore (DINO-v2) \cite{PatchCore} & RGB 
& \underline{0.790} & \underline{0.489} & \textbf{0.792} & 0.609 & \textbf{0.715} & \textbf{0.756} & \textbf{0.290}& \underline{0.459} & \underline{0.613} \\

EfficientAD \cite{Efficient_AD} & RGB 
& 0.682 & 0.462 & 0.763 & 0.534 & 0.680 & 0.743 & 0.407 & \textbf{0.488} & 0.595 \\

AST \cite{AST} & RGB 
& 0.595 & 0.354 & 0.754 & 0.338 & 0.594 & 0.748 & 0.231 & 0.083 & 0.462 \\

MVAD \cite{MVAD} & RGB 
& 0.754 & 0.481 & 0.760 & 0.359 & 0.655 & 0.745 & 0.256 & 0.326 & 0.542 \\

BTF \cite{BTF} & RGB + PC 
& 0.551 & 0.402 & 0.750 & 0.377 & 0.614 & 0.741 & 0.092 & 0.030 & 0.445 \\

CFM (DINO-v2+FPFH) \cite{CFM} & RGB + PC 
& 0.611 & 0.432 & 0.770 & 0.449 & 0.638 & 0.743 & 0.117 & 0.166 & 0.491 \\

M3DM (DINO-v2+FPFH) \cite{M3DM} & RGB + PC 
& 0.733 & 0.452 & 0.767 & \textbf{0.702} & 0.702 & 0.752 & 0.288 & 0.126 & 0.565 \\

AST \cite{AST} & RGB + Depth 
& 0.723 & 0.443 & 0.765 & 0.313 & 0.683 & 0.749 & 0.164 & 0.166 & 0.501 \\

\textbf{SGANet (Ours)} & RGB + Depth 
& \textbf{0.798} & \textbf{0.504} & \underline{0.777} & \underline{0.615} & \textbf{0.715} & \underline{0.754} & \underline{0.289} & 0.456 & \textbf{0.614} \\
\hline 
\end{tabular}
}
\end{table*}

\textbf{Quantitative Results on Eyecandies.}
Tables \ref{tab:Eyecandies I-AUROC}, \ref{tab:Eyecandies P-AUROC}, and \ref{tab:Eyecandies AUPRO@30 percent} present the results of I-AUROC, P-AUROC, and AUPRO@30\% on the Eyecandies dataset, respectively. SGANet achieves the best performance across all three metrics, demonstrating its effectiveness for multimodal multi-view anomaly detection.
For anomaly detection, SGANet obtains a mean I-AUROC of 0.743, outperforming all other baselines. The improvement is particularly notable in categories such as Chocolate Praline, Lollipop, and Peppermint Candy, where structural cues from multiple viewpoints facilitate the detection of subtle defects.
For anomaly localization, SGANet achieves the highest mean P-AUROC of 0.854 and AUPRO@30\% of 0.555. 
This improvement stems from the joint modeling of multimodal and multi-view interactions in SGANet, where complementary appearance and geometric cues are effectively integrated across viewpoints, leading to more precise and spatially consistent localization.

\begin{figure}
	\centering
	\includegraphics[width=\columnwidth]{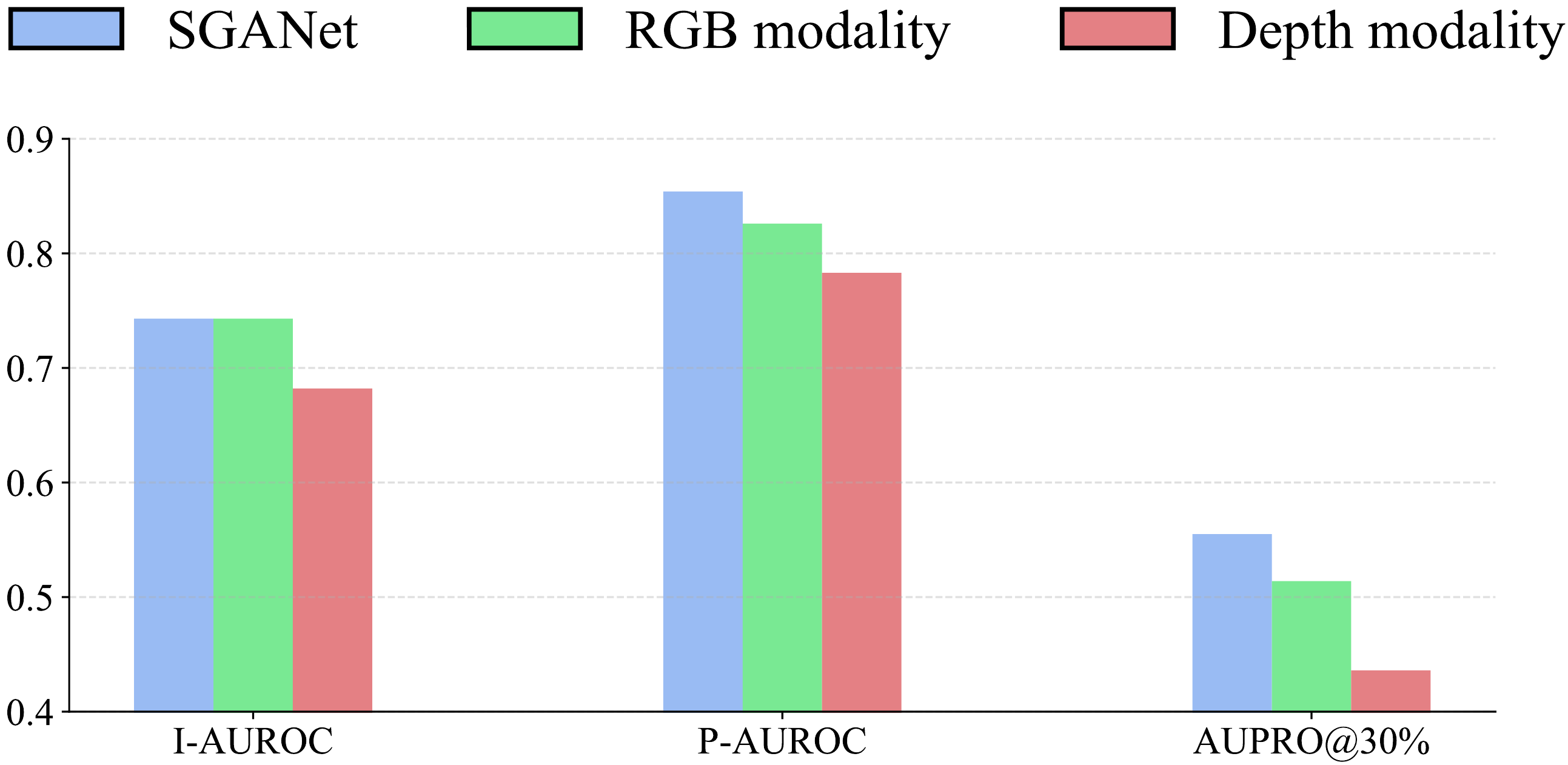}
	\caption{I-AUROC, P-AUROC and AUPRO@30\% results on the Eyecandies dataset using RGB modality, depth modality, and the proposed SGANet framework.}
	\label{Fig: Eyecandies Histogram}
\end{figure}

Fig. \ref{Fig: Eyecandies Histogram} further compares unimodal and multimodal configurations. The multimodal configuration achieves the best overall performance,  
with mean I-AUROC, P-AUROC, and AUPRO@30\% of 0.743, 0.854, and 0.555, respectively, compared with 0.743, 0.826, and 0.514 for the RGB modality and 0.682, 0.783, and 0.436 for the Depth modality.
While the I-AUROC remains the same compared to the RGB modality, noticeable gains are observed in P-AUROC and AUPRO@30\%, indicating that depth information primarily contributes to accurate spatial localization. 
Depth alone provides limited discriminative capability. 
However, when combined with RGB, it enables the integration of complementary structural information, leading to improved localization accuracy and overall robustness, as geometric information from depth complements appearance cues and reduces spurious responses in normal regions.

\begin{table*}[ht]
\centering
\caption{I-AUROC result on the Eyecandies dataset. Best results in bold, runner-ups \underline{underlined}.}
\label{tab:Eyecandies I-AUROC}
\resizebox{\textwidth}{!}{
\begin{tabular}{ccccccccccccc}
\hline 
Category & Modality & Candy. Cane & Choco. Cook. & Choco. Pra. & Conf. & Gum. Bear & Hazel. Truf. & Lico. Sand. & Lolli. & Marsh. & Pep. Candy & Mean \\
\hline
PatchCore (WRN-101) \cite{PatchCore} & RGB               
& 0.381 & 0.592 & 0.534 & 0.677 & 0.409 & \textbf{0.707} & 0.448 & 0.621 & 0.651 & 0.562 & 0.558 \\

PatchCore (DINO-v2) \cite{PatchCore} & RGB               
& 0.443 & \underline{0.824} & 0.563 & 0.680 & \textbf{0.784} & \underline{0.602} & \textbf{0.811} & 0.658 & \underline{0.781} & 0.587 & \underline{0.673} \\

MVAD \cite{MVAD}  & RGB 
& 0.322 & 0.534 & 0.448 & \textbf{0.773} & 0.579 & 0.410 & 0.635 & 0.487 & 0.666 & \underline{0.723} & 0.558 \\

CFM (DINO-v2+FPFH) \cite{CFM} & RGB + PC 
& 0.390 & 0.542 & 0.514 & 0.643 & 0.572 & 0.426 & 0.613 & 0.618 & 0.688 & 0.533 & 0.554 \\

AST \cite{AST} & RGB + Depth       
& \underline{0.448} & 0.424 & \underline{0.574} & 0.370 & 0.548 & 0.595 & 0.613 & \underline{0.662} & 0.498 & 0.325 & 0.506         \\

SGANet (Ours) & RGB + Depth       \
& \textbf{0.451} & \textbf{0.925} & \textbf{0.787} & \underline{0.757} & \underline{0.692} & 0.534 & \underline{0.795} & \textbf{0.770}    & \textbf{0.842} & \textbf{0.874} & \textbf{0.743} \\  
\hline 
\end{tabular}
}
\end{table*}

\begin{table*}[ht]
\centering
\caption{P-AUROC result on the Eyecandies dataset. Best results in bold, runner-ups \underline{underlined}.}
\label{tab:Eyecandies P-AUROC}
\resizebox{\textwidth}{!}{
\begin{tabular}{ccccccccccccc}
\hline
Category & Modality & Candy. Cane & Choco. Cook. & Choco. Pra. & Conf. & Gum. Bear & Hazel. Truf. & Lico. Sand. & Lolli. & Marsh. & Pep. Candy & Mean \\
\hline
PatchCore (WRN-101) \cite{PatchCore} & RGB 
& 0.911 & 0.823 & 0.651 & 0.787 & 0.777 & \underline{0.712} & 0.847 &\textbf{0.960} &\textbf{0.891} & 0.772 &0.813 \\

PatchCore (DINO-v2) \cite{PatchCore} & RGB         
& 0.601 & \underline{0.908} &\underline{0.733} & \underline{0.872} & \underline{0.873} & \underline{0.712} & \textbf{0.893} 
& 0.816 & \underline{0.888} & \underline{0.821} & 0.812 \\

MVAD \cite{MVAD} & RGB 
& \textbf{0.933} & 0.847 & 0.685 & 0.814 & 0.855 &\textbf{0.736} & 0.855 & \underline{0.956} & 0.842 & 0.773 & \underline{0.830} \\

CFM (DINO-v2 + FPFH) \cite{CFM} & RGB +PC 
&\underline{0.928} & 0.775 & 0.706 & 0.764 & 0.797 &0.654 & 0.876 &0.946 & 0.871 & 0.802 & 0.812 \\

AST \cite{AST} & RGB + Depth 
& 0.750 & 0.647 & 0.566 & 0.662 & 0.559 & 0.519 & 0.572 & 0.587 & 0.479 & 0.471 & 0.581 \\

SGANet (Ours) & RGB + Depth 
& 0.846 & \textbf{0.921} & \textbf{0.795} & \textbf{0.879} &\textbf{0.899} & \underline{0.712} &\underline{0.887} & 0.885 & 0.884 &\textbf{0.834} &\textbf{0.854}\\
\hline 
\end{tabular}
}
\end{table*}

\begin{table*}[h]
\centering
\caption{AUPRO@30\% result on the Eyecandies dataset. Best results in bold, runner-ups \underline{underlined}.}
\label{tab:Eyecandies AUPRO@30 percent}
\resizebox{\textwidth}{!}{
\begin{tabular}{ccccccccccccc}
\hline
Category & Modality          & Candy. Cane & Choco. Cook. & Choco. Pra. & Conf. & Gum. Bear & Hazel. Truf. & Lico. Sand. & Lolli. & Marsh. & Pep. Candy & Mean \\
\hline
PatchCore (WRN-101) \cite{PatchCore} & RGB         
&0.665 & 0.497 & 0.174 & 0.369 & 0.305 &0.276 & 0.471 & \textbf{0.771} &0.565 & 0.395 & 0.449 \\

PatchCore (DINO-v2) \cite{PatchCore} & RGB & 0.177 &\underline{0.715} &\underline{0.325} & \underline{0.660} & \textbf{0.614} & \textbf{0.322} & \textbf{0.695} & 0.328 & \textbf{0.626} & \underline{0.491} &\underline{0.495} \\

MVAD \cite{MVAD} & RGB 
& \underline{0.717} & 0.508 & 0.196 & 0.489 & 0.388 & 0.210 & 0.454 & \underline{0.770} & 0.523 & 0.399 & 0.465\\

CFM (DINO-v2 + FPFH) \cite{CFM} & RGB +PC 
& \textbf{0.742} & 0.395 & 0.278 & 0.387 & 0.286 &0.185 & 0.544 &0.744 & 0.524 & 0.405 & 0.449 \\

AST \cite{AST} & RGB + Depth 
& 0.431 & 0.293 & 0.202 & 0.242 & 0.214 & 0.130 & 0.130 & 0.163 & 0.041 & 0.021 & 0.187 \\

SGANet (Ours) & RGB + Depth 
& 0.490 & \textbf{0.750} & \textbf{0.520} & \textbf{0.665} &\underline{0.591} &\underline{0.310} &\underline{0.645} & 0.470 
& \underline{0.583} & \textbf{0.527} & \textbf{0.555}   \\
\hline 
\end{tabular}
}
\end{table*}

\begin{table}[ht]
\centering
\caption{Parameter analysis of hyperparameter $k$ on the SiM3D dataset.}
\label{tab:Ablation Study of k on SiM3D}
\fontsize{10pt}{12pt}\selectfont
\begin{tabular}{ccc}
\hline 
\multicolumn{1}{l}{} & I-AUROC & V-AUPRO@1\% \\\hline  
$k = 2 $            & 0.872   & \textbf{0.614}       \\
$k = 4 $            & 0.873   & 0.612       \\
$k = 6 $           & 0.876   & 0.613       \\
$k = 8 $            & \textbf{0.887}   & \textbf{0.614}       \\
$k = 10 $          & 0.885   & 0.613       \\
\hline 
\end{tabular}
\end{table}

\begin{table}[ht]
\centering
\caption{Parameter analysis of the number of neighboring views $N$ on the SiM3D dataset.}
\label{tab:Ablation Study of N on SiM3D}
\fontsize{10pt}{12pt}\selectfont
\begin{tabular}{ccc}
\hline 
\multicolumn{1}{l}{} & I-AUROC & V-AUPRO@1\% \\\hline  
$N=1$         & 0.871   & 0.612       \\
$N=2$         & \textbf{0.887}   & \textbf{0.614}       \\
$N=3$        & 0.881   & \textbf{0.614}       \\
$N=4$         & 0.871   & 0.612       \\
\hline 
\end{tabular}
\end{table}

\begin{table}[ht]
\centering
% \small
% \setlength{\tabcolsep}{3pt} % 默认6pt
\caption{Ablation study of loss components on the SiM3D and Eyecandies datasets.}
\label{tab:Ablation Study of Loss functions on both datasets}
\fontsize{10pt}{12pt}\selectfont
\begin{tabular}{cccccc}
\hline
& $\mathcal{L}_{\text{SSPA}}$ & \ding{52} & \ding{52} & \ding{52} & \ding{52} \\ 
& $\mathcal{L}_{\text{view}}$ & \ding{55} & \ding{52} & \ding{55} & \ding{52} \\
& $\mathcal{L}_{\text{diff}}$ & \ding{55} & \ding{55} & \ding{52} & \ding{52} \\ \hline

\multirow{2}{*}{SiM3D} 
& I-AUROC     & 0.870 & 0.874 & 0.874 & \textbf{0.887} \\
& V-AUPRO@1\% & 0.612 & 0.612 & 0.613 & \textbf{0.614} \\ \hline

\multirow{3}{*}{Eyecandies}
& I-AUROC     & 0.732 & 0.740 & 0.737 & \textbf{0.743} \\
& P-AUROC     & 0.845 & 0.851 & 0.853 & \textbf{0.854} \\ 
& AUPRO@30\%  & 0.538 & 0.551 & \textbf{0.555} & \textbf{0.555} \\ 
\hline 
\end{tabular}
\end{table}

\begin{figure*}
	\centering
	\includegraphics[width=1.\textwidth]{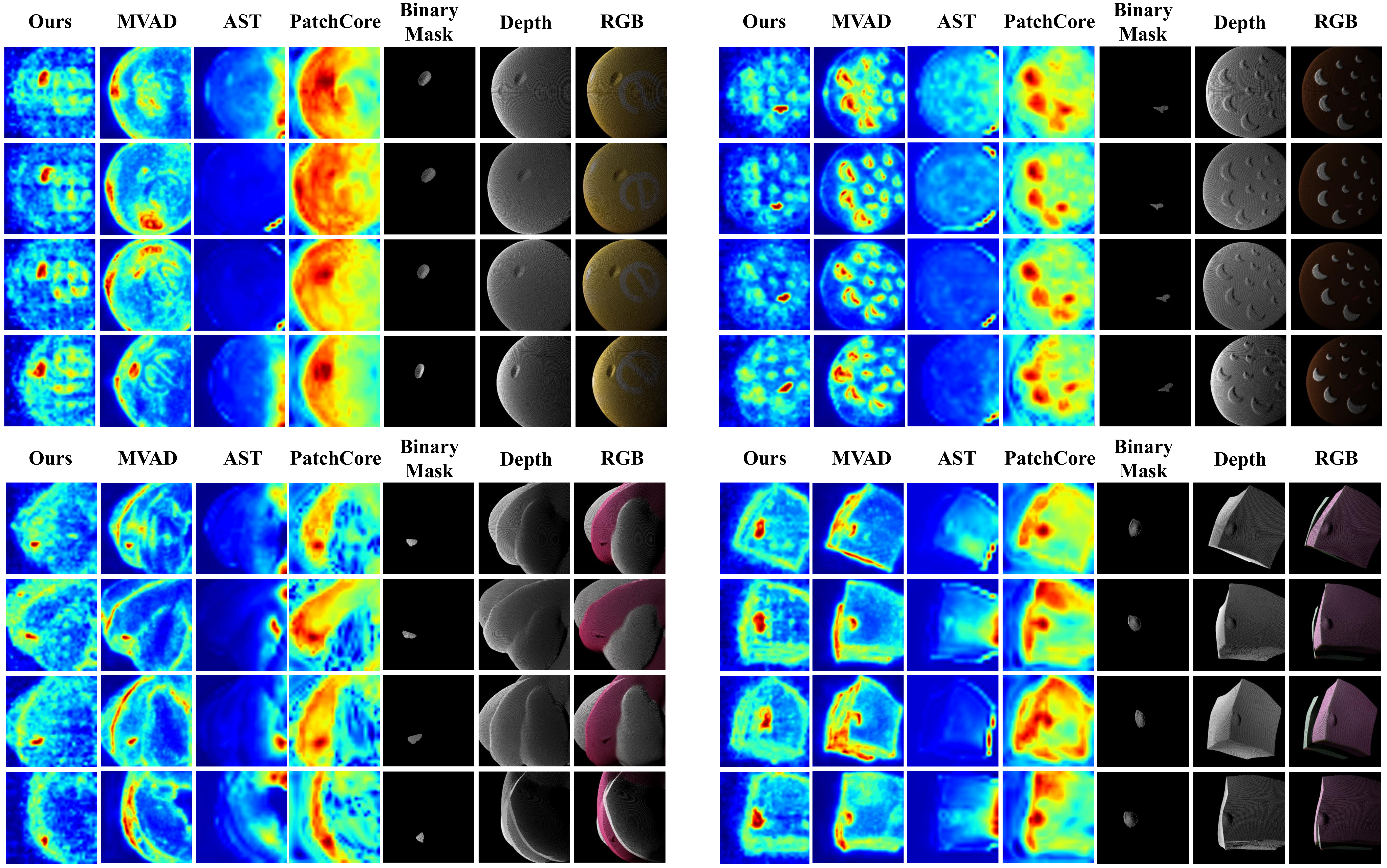}
	\caption{Qualitative comparison on the Eyecandies dataset \cite{Eyecandies}. Visualizations include RGB images (RGB), depth maps (Depth), binary masks (Binary Mask), and the corresponding anomaly score maps predicted by representative methods, including single-view baselines (PatchCore, AST), the multi-view method (MVAD), and the proposed SGANet framework.}
	\label{Fig: Eyecandies Visulization Result}
\end{figure*}

\subsection{Ablation Studies}

\textbf{Analysis of the Number of Selected Patches $k$ in SCFRM.}
Table \ref{tab:Ablation Study of k on SiM3D} shows the effect of varying the number of selected cross-view patches $k$. As $k$ increases from 2 to 8, the I-AUROC improves from 0.872 to 0.887, indicating that incorporating additional informative cross-view patches benefits representation learning, while V-AUPRO@1\% remains relatively stable. 
When $k$ further increases to 10, the I-AUROC slightly decreases, suggesting that excessive patch aggregation introduces noisy or irrelevant features that weaken anomaly discrimination.

\textbf{Analysis of the Number of Neighboring Views $N$.}
Table \ref{tab:Ablation Study of N on SiM3D} reports the effect of varying the number of neighboring views used for geometric correspondence alignment.
As shown in Table \ref{tab:Ablation Study of N on SiM3D}, the best performance is achieved at $N=2$. Using only one neighboring view provides limited cross-view information and leads to weaker alignment, while increasing $N$ to 4 introduces less correlated views and slightly degrades performance. This indicates that a compact neighborhood is more effective for enforcing geometric consistency.

\textbf{Analysis of Individual Loss Components.}
Table \ref{tab:Ablation Study of Loss functions on both datasets} reports the effect of different loss terms in our framework on both the SiM3D and Eyecandies datasets.
Using $\mathcal{L}_{\text{SSPA}}$ as the baseline for metric learning, adding either $\mathcal{L}_{\text{view}}$ or $\mathcal{L}_{\text{diff}}$ improves performance, and jointly optimizing both losses achieves the best results on both datasets.
Specifically, $\mathcal{L}_{\text{view}}$ enforces semantic alignment between image and depth features at corresponding spatial locations, thereby reducing cross-modal feature discrepancies. 
Meanwhile, $\mathcal{L}_{\text{diff}}$ promotes structural consistency of differential features across viewpoints, encouraging the model to capture intrinsic structural variations rather than changes caused by viewpoint transformations. 
By jointly modeling semantic correspondence and structural consistency across viewpoints, the two losses provide complementary supervision, leading to improved performance in both anomaly detection and localization tasks.

\section{Conclusion}
In this paper, we propose SGANet, a unified framework for multimodal multi-view anomaly detection that integrates semantic and geometric alignment to learn physically consistent feature representations across different viewpoints and modalities.
SGANet enforces multiple forms of feature consistency through three key components: the Selective Cross-View Feature Refinement Module (SCFRM), the Semantic-Structural Patch Alignment (SSPA), and the Multi-View Geometric Alignment (MVGA), enabling unified feature representation learning.
By explicitly modeling feature correspondences across viewpoints and modalities, SGANet learns geometrically coherent representations for normal regions, thereby enhancing anomaly detection performance.
Extensive experiments on the SiM3D and Eyecandies datasets demonstrate that SGANet achieves state-of-the-art performance in both anomaly detection and localization tasks, validating the effectiveness and generalizability of the proposed framework.

\bibliographystyle{elsarticle-num} 
\bibliography{reference}

@InProceedings{CFM,
    author    = {Costanzino, Alex and Ramirez, Pierluigi Zama and Lisanti, Giuseppe and Di Stefano, Luigi},
    title     = {Multimodal Industrial Anomaly Detection by Crossmodal Feature Mapping},
    booktitle = {Proceedings of the IEEE/CVF Conference on Computer Vision and Pattern Recognition (CVPR)},
    month     = {June},
    year      = {2024},
    pages     = {17234-17243}
}

@ARTICLE{2M3DF,
  author={Asad, Mujtaba and Azeem, Waqar and Jiang, He and Tayyab Mustafa, Hafiz and Yang, Jie and Liu, Wei},
  journal={IEEE Transactions on Circuits and Systems for Video Technology}, 
  title={2M3DF: Advancing 3D Industrial Defect Detection With Multi-Perspective Multimodal Fusion Network}, 
  year={2025},
  volume={35},
  number={7},
  pages={6803-6815}}

@InProceedings{M3DM,
    author    = {Wang, Yue and Peng, Jinlong and Zhang, Jiangning and Yi, Ran and Wang, Yabiao and Wang, Chengjie},
    title     = {Multimodal Industrial Anomaly Detection via Hybrid Fusion},
    booktitle = {Proceedings of the IEEE/CVF Conference on Computer Vision and Pattern Recognition (CVPR)},
    month     = {June},
    year      = {2023},
    pages     = {8032-8041}
}

@InProceedings{BTF,
    author    = {Horwitz, Eliahu and Hoshen, Yedid},
    title     = {Back to the Feature: Classical 3D Features Are (Almost) All You Need for 3D Anomaly Detection},
    booktitle = {Proceedings of the IEEE/CVF Conference on Computer Vision and Pattern Recognition (CVPR) Workshops},
    month     = {June},
    year      = {2023},
    pages     = {2968-2977}
}

@InProceedings{Shape-Guided,
  title = {Shape-Guided Dual-Memory Learning for 3D Anomaly Detection},
  author = {Chu, Yu-Min and Liu, Chieh and Hsieh, Ting-I and Chen, Hwann-Tzong and Liu, Tyng-Luh},
  booktitle = {Proceedings of the 40th International Conference on Machine Learning},
  pages = {6185--6194},
  year = {2023},
}

@InProceedings{G2SF,
    author    = {Tao, Chengyu and Cao, Xuanming and Du, Juan},
    title     = {G2SF: Geometry-Guided Score Fusion for Multimodal Industrial Anomaly Detection},
    booktitle = {Proceedings of the IEEE/CVF International Conference on Computer Vision (ICCV)},
    month     = {October},
    year      = {2025},
    pages     = {20551-20560}
}

@article{CPMF,
title = {Complementary pseudo multimodal feature for point cloud anomaly detection},
journal = {Pattern Recognition},
volume = {156},
pages = {110761},
year = {2024},
issn = {0031-3203},
author = {Yunkang Cao and Xiaohao Xu and Weiming Shen},
}

@article{MVAD,
  title={Learning Multi-view Anomaly Detection},
  author={He, Haoyang and Zhang, Jiangning and Tian, Guanzhong and Wang, Chengjie and Xie, Lei},
  journal={arXiv preprint arXiv:2407.11935},
  year={2024}
}

@misc{MVEAD,
  title={Multi-View Industrial Anomaly Detection with Epipolar Constrained Cross-View Fusion}, 
  author={Yifan Liu and Xun Xu and Shijie Li and Jingyi Liao and Xulei Yang},
  year={2025},
  eprint={2503.11088},
  archivePrefix={arXiv},
  primaryClass={cs.CV} 
}

@InProceedings{PatchCore,
    author    = {Roth, Karsten and Pemula, Latha and Zepeda, Joaquin and Sch\"olkopf, Bernhard and Brox, Thomas and Gehler, Peter},
    title     = {Towards Total Recall in Industrial Anomaly Detection},
    booktitle = {Proceedings of the IEEE/CVF Conference on Computer Vision and Pattern Recognition (CVPR)},
    month     = {June},
    year      = {2022},
    pages     = {14318-14328}
}

@ARTICLE{A_Systematic_Exploration-TNNLS,
  author={Wang, Siqi and Liu, Jiyuan and Yu, Guang and Liu, Xinwang and Zhou, Sihang and Zhu, En and Yang, Yuexiang and Yin, Jianping and Yang, Wenjing},
  journal={IEEE Transactions on Neural Networks and Learning Systems}, 
  title={Multiview Deep Anomaly Detection: A Systematic Exploration}, 
  year={2024},
  volume={35},
  number={2},
  pages={1651-1665}}

@inproceedings{SiM3D,
  author    = {Costanzino, Alex and Zama Ramirez, Pierluigi and Lella, Luigi and Ragaglia, Matteo and Oliva, Alessandro and Lisanti, Giuseppe and Di Stefano, Luigi},
  title     = {SiM3D: Single-instance Multiview Multimodal and Multisetup 3D Anomaly Detection Benchmark},
  booktitle = {International Conference on Computer Vision (ICCV)},
  year      = {2025},
}

@InProceedings{Eyecandies,
    author    = {Bonfiglioli, Luca and Toschi, Marco and Silvestri, Davide and Fioraio, Nicola and De Gregorio, Daniele},
    title     = {The Eyecandies Dataset for Unsupervised Multimodal Anomaly Detection and Localization},
    booktitle = {Proceedings of the Asian Conference on Computer Vision (ACCV)},
    month     = {December},
    year      = {2022},
    pages     = {3586-3602}
}

@misc{Dino-v2,
      title={DINOv2: Learning Robust Visual Features without Supervision}, 
      author={Maxime Oquab and Timothée Darcet and Théo Moutakanni and Huy Vo and Marc Szafraniec and Vasil Khalidov and Pierre Fernandez and Daniel Haziza and Francisco Massa and Alaaeldin El-Nouby and Mahmoud Assran and Nicolas Ballas and Wojciech Galuba and Russell Howes and Po-Yao Huang and Shang-Wen Li and Ishan Misra and Michael Rabbat and Vasu Sharma and Gabriel Synnaeve and Hu Xu and Hervé Jegou and Julien Mairal and Patrick Labatut and Armand Joulin and Piotr Bojanowski},
      year={2024},
      eprint={2304.07193},
      archivePrefix={arXiv},
      primaryClass={cs.CV}
}

@InProceedings{Efficient_AD,
    author    = {Batzner, Kilian and Heckler, Lars and K\"onig, Rebecca},
    title     = {EfficientAD: Accurate Visual Anomaly Detection at Millisecond-Level Latencies},
    booktitle = {Proceedings of the IEEE/CVF Winter Conference on Applications of Computer Vision (WACV)},
    month     = {January},
    year      = {2024},
    pages     = {128-138}
}

@InProceedings{AST,
    author    = {Rudolph, Marco and Wehrbein, Tom and Rosenhahn, Bodo and Wandt, Bastian},
    title     = {Asymmetric Student-Teacher Networks for Industrial Anomaly Detection},
    booktitle = {Proceedings of the IEEE/CVF Winter Conference on Applications of Computer Vision (WACV)},
    month     = {January},
    year      = {2023},
    pages     = {2592-2602}
}

@misc{cao2024surveyvisualanomalydetection,
title={A Survey on Visual Anomaly Detection: Challenge, Approach, and Prospect}, 
author={Yunkang Cao and Xiaohao Xu and Jiangning Zhang and Yuqi Cheng and Xiaonan Huang and Guansong Pang and Weiming Shen},
year={2024},
eprint={2401.16402},
archivePrefix={arXiv},
primaryClass={cs.CV}
}

@misc{liang20253d,
  title={3d anomaly detection: A survey},
  author={Liang, HANZHE and Guo, BINGYANG and Huang, YAWEN and Lyu, JIAYI and Gao, CAN and Cao, YUNKANG and Wang, JINBAO and Yu, RUIYUN and Shen, LINLIN and Li, PAN},
  year={2025},
  publisher={ArXiv}
}

@ARTICLE{9810850,
  author={Wang, Siqi and Liu, Jiyuan and Yu, Guang and Liu, Xinwang and Zhou, Sihang and Zhu, En and Yang, Yuexiang and Yin, Jianping and Yang, Wenjing},
  journal={IEEE Transactions on Neural Networks and Learning Systems}, 
  title={Multiview Deep Anomaly Detection: A Systematic Exploration}, 
  year={2024},
  volume={35},
  number={2},
  pages={1651-1665}}

@article{IAENet,
title = {IAENet: An importance-aware ensemble model for 3D point cloud-based anomaly detection},
journal = {Information Fusion},
volume = {130},
pages = {104097},
year = {2026},
issn = {1566-2535},
author = {Xuanming Cao and Chengyu Tao and Yifeng Cheng and Juan Du},
}

@InProceedings{3D-ST,
    author    = {Bergmann, Paul and Sattlegger, David},
    title     = {Anomaly Detection in 3D Point Clouds Using Deep Geometric Descriptors},
    booktitle = {Proceedings of the IEEE/CVF Winter Conference on Applications of Computer Vision (WACV)},
    month     = {January},
    year      = {2023},
    pages     = {2613-2623}
}

@article{LI2025103356,
title = {A multi-scale information fusion framework with interaction-aware global attention for industrial vision anomaly detection and localization},
journal = {Information Fusion},
volume = {124},
pages = {103356},
year = {2025},
issn = {1566-2535},
author = {Zhuo Li and Yifei Ge and Lin Meng},
}

@inproceedings{PaDiM,
author = {Defard, Thomas and Setkov, Aleksandr and Loesch, Angelique and Audigier, Romaric},
title = {PaDiM: A Patch Distribution Modeling Framework for Anomaly Detection and Localization},
year = {2021},
isbn = {978-3-030-68798-4},
publisher = {Springer-Verlag},
address = {Berlin, Heidelberg},
booktitle = {Pattern Recognition. ICPR International Workshops and Challenges: Virtual Event, January 10–15, 2021, Proceedings, Part IV},
pages = {475–489},
numpages = {15},
}

@misc{yu2025learningmultiviewmulticlassanomaly,
      title={Learning Multi-view Multi-class Anomaly Detection}, 
      author={Qianzi Yu and Yang Cao and Yu Kang},
      year={2025},
      eprint={2504.21294},
      archivePrefix={arXiv},
      primaryClass={cs.CV}
}

@article{Mao_Lian_Wang_Liu_Zheng_Wei_2025, 
title={Unveiling Multi-View Anomaly Detection: Intra-view Decoupling and Inter-view Fusion}, 
volume={39},
number={12}, 
journal={Proceedings of the AAAI Conference on Artificial Intelligence}, 
author={Mao, Kai and Lian, Yiyang and Wang, Yangyang and Liu, Meiqin and Zheng, Nanning and Wei, Ping}, 
year={2025}, 
month={Apr.}, 
pages={12381-12389} }

@INPROCEEDINGS{DRAEM,
  author={Zavrtanik, Vitjan and Kristan, Matej and Skočaj, Danijel},
  booktitle={2021 IEEE/CVF International Conference on Computer Vision (ICCV)}, 
  title={DRÆM – A discriminatively trained reconstruction embedding for surface anomaly detection}, 
  year={2021},
  volume={},
  number={},
  pages={8310-8319}}

@InProceedings{CFLOW-AD,
    author    = {Gudovskiy, Denis and Ishizaka, Shun and Kozuka, Kazuki},
    title     = {CFLOW-AD: Real-Time Unsupervised Anomaly Detection With Localization via Conditional Normalizing Flows},
    booktitle = {Proceedings of the IEEE/CVF Winter Conference on Applications of Computer Vision (WACV)},
    month     = {January},
    year      = {2022},
    pages     = {98-107}
}

@InProceedings{PNI,
    author    = {Bae, Jaehyeok and Lee, Jae-Han and Kim, Seyun},
    title     = {PNI : Industrial Anomaly Detection using Position and Neighborhood Information},
    booktitle = {Proceedings of the IEEE/CVF International Conference on Computer Vision (ICCV)},
    month     = {October},
    year      = {2023},
    pages     = {6373-6383}
}

@inproceedings{DSR,
    author = {Zavrtanik, Vitjan and Kristan, Matej and Sko\v{c}aj, Danijel},
    title = {DSR – A Dual Subspace Re-Projection Network for Surface Anomaly Detection},
    year = {2022},
    isbn = {978-3-031-19820-5},
    publisher = {Springer-Verlag},
    address = {Berlin, Heidelberg},
    booktitle = {Computer Vision – ECCV 2022: 17th European Conference, Tel Aviv, Israel, October 23–27, 2022, Proceedings, Part XXXI},
    pages = {539–554},
    numpages = {16},
    location = {Tel Aviv, Israel}
}

@inproceedings{Anomalydiffusion,
  title={AnomalyDiffusion: Few-Shot Anomaly Image Generation with Diffusion Model},
  author={Hu, Teng and Zhang, Jiangning and Yi, Ran and Du, Yuzhen and Chen, Xu and Liu, Liang and Wang, Yabiao and Wang, Chengjie},
  booktitle={Proceedings of the AAAI Conference on Artificial Intelligence},
  year={2024}
}

@inproceedings{LSFA,
  title={Self-supervised feature adaptation for 3d industrial anomaly detection},
  author={Tu, Yuanpeng and Zhang, Boshen and Liu, Liang and Li, Yuxi and Zhang, Jiangning and Wang, Yabiao and Wang, Chengjie and Zhao, Cairong},
  booktitle={European Conference on Computer Vision},
  pages={75--91},
  year={2024},
  organization={Springer}
}

@article{InfoNCE,
  title={Representation learning with contrastive predictive coding},
  author={Oord, Aaron van den and Li, Yazhe and Vinyals, Oriol},
  journal={arXiv preprint arXiv:1807.03748},
  year={2018}
}

@article{FANG2020107474,
title = {A novel hybrid approach for crack detection},
journal = {Pattern Recognition},
volume = {107},
pages = {107474},
year = {2020},
issn = {0031-3203},
author = {Fen Fang and Liyuan Li and Ying Gu and Hongyuan Zhu and Joo-Hwee Lim},
}

@article{YANG2022108874,
title = {Learning deep feature correspondence for unsupervised anomaly detection and segmentation},
journal = {Pattern Recognition},
volume = {132},
pages = {108874},
year = {2022},
issn = {0031-3203},
author = {Jie Yang and Yong Shi and Zhiquan Qi},
}

@inproceedings{PointAD,
 author = {Zhou, Qihang and Yan, Jiangtao and He, Shibo and Meng, Wenchao and Chen, Jiming},
 booktitle = {Advances in Neural Information Processing Systems},
 editor = {A. Globerson and L. Mackey and D. Belgrave and A. Fan and U. Paquet and J. Tomczak and C. Zhang},
 pages = {84866--84896},
 publisher = {Curran Associates, Inc.},
 title = {PointAD: Comprehending 3D Anomalies from Points and Pixels for Zero-shot 3D Anomaly Detection},
 volume = {37},
 year = {2024}
}

@article{ZAVRTANIK2021107706,
title = {Reconstruction by inpainting for visual anomaly detection},
journal = {Pattern Recognition},
volume = {112},
pages = {107706},
year = {2021},
issn = {0031-3203},
author = {Vitjan Zavrtanik and Matej Kristan and Danijel Skočaj},
}

@article{NIE2026113261,
title = {Few-shot medical anomaly detection through centroid consultation back and test-time self-calibration},
journal = {Pattern Recognition},
volume = {178},
pages = {113261},
year = {2026},
issn = {0031-3203},
author = {Zihan Nie and Muhao Xu and Yuan Cui and Hua Wei and Wei Yi and Sijie Niu and Yi Wan and Xunbin Wei and Weiye Song},
}

@article{MMRD, 
title={Rethinking Reverse Distillation for Multi-Modal Anomaly Detection}, 
volume={38}, 
number={8}, 
journal={Proceedings of the AAAI Conference on Artificial Intelligence}, 
author={Gu, Zhihao and Zhang, Jiangning and Liu, Liang and Chen, Xu and Peng, Jinlong and Gan, Zhenye and Jiang, Guannan and Shu, Annan and Wang, Yabiao and Ma, Lizhuang}, 
year={2024}, 
month={Mar.}, 
pages={8445-8453} }

@ARTICLE{M3DM-NR,
  author={Wang, Chengjie and Zhu, Haokun and Peng, Jinlong and Wang, Yue and Yi, Ran and Wu, Yunsheng and Ma, Lizhuang and Zhang, Jiangning},
  journal={IEEE Transactions on Pattern Analysis and Machine Intelligence}, 
  title={M3DM-NR: RGB-3D Noisy-Resistant Industrial Anomaly Detection via Multimodal Denoising}, 
  year={2025},
  volume={47},
  number={11},
  pages={9981-9993}}

@article{du20253d,
  title={3D vision-based anomaly detection in manufacturing: A survey},
  author={Du, Juan and Tao, Chengyu and Cao, Xuanming and Tsung, Fugee},
  journal={Frontiers of Engineering Management},
  volume={12},
  number={2},
  pages={343--360},
  year={2025},
  publisher={Springer}
}

@article{VSAD-AAAI, 
title={Unsupervised Multi-View Visual Anomaly Detection via Progressive Homography-Guided Alignment}, 
volume={40}, 
number={4}, 
journal={Proceedings of the AAAI Conference on Artificial Intelligence}, 
author={Chen, Xintao and Xu, Xiaohao and Zheng, Bozhong and Liu, Yun and Wu, Yingna}, 
year={2026}, 
month={Mar.}, 
pages={3065-3073} }




% %% For numbered reference style

%   \textit{\LaTeX: a document preparation system},
%   Addison Wesley, Massachusetts,

% \end{thebibliography}
\end{document}